\pdfoutput=1

\documentclass[11pt]{article}
\usepackage{arabtex}
\usepackage{utf8}
\usepackage{float}
\usepackage{multirow}
\setcode{utf8}

\usepackage[final]{acl}

\usepackage{times}
\usepackage{latexsym}

\usepackage[T1]{fontenc}

\usepackage[utf8]{inputenc}

\usepackage{microtype}

\usepackage{xcolor}

\usepackage{inconsolata}

\usepackage{graphicx}

\title{TEDxTN: A Three-way Speech Translation Corpus for Code-Switched Tunisian Arabic - English}

\author{
 \textbf{Fethi Bougares\textsuperscript{1,2}},
 \textbf{Salima Mdhaffar\textsuperscript{2}},
 \textbf{Haroun Elleuch\textsuperscript{1,2}},
 \textbf{Yannick Estève\textsuperscript{2}}
\\
 \textsuperscript{1}ELYADATA, Paris, France,
 \textsuperscript{2}Laboratoire Informatique d'Avignon, Avignon, France
\\
 \small{
   \textbf{Correspondence:} \href{mailto:email@domain}{fethi.bougares@elyadata.com}
 }
}

\begin{document}
\maketitle
\begin{abstract}

In this paper, we introduce TEDxTN, the first publicly available Tunisian Arabic to English speech translation dataset. This work is in line with the ongoing effort to mitigate the data scarcity obstacle for a number of Arabic dialects. We collected, segmented, transcribed and translated 108 TEDx talks following our internally developed annotations guidelines. The collected talks represent 25 hours of speech with code-switching that cover speakers with various accents from over 11 different regions of Tunisia. We make the annotation guidelines and corpus publicly available. This will enable the extension of TEDxTN to new talks as they become available. We also report results for strong baseline systems of Speech Recognition and Speech Translation using multiple pre-trained and fine-tuned end-to-end models. This corpus is the first open source and publicly available speech translation corpus of Code-Switching Tunisian dialect. We believe that this is a valuable resource that can motivate and facilitate further research on the natural language processing of Tunisian Dialect.

\end{abstract}

\section{Introduction}

\label{sec:intro}

Speech translation is the task of translating speech in a given source language into text in another target language. This task is traditionally accomplished through a cascading approach, where a first Automatic Speech Recognition system (ASR) recognizes spoken words, followed by a Machine Translation (MT) system that translates the transcribed text into the target language. This approach is generally criticized because it suffers from cascaded error propagation and high resource and training costs \citep{Sethiya2025}. To overcome these weaknesses, researchers proposed end-to-end (E2E) models \citep{cho2014} \citep{bahdanau2016} \citep{vaswani2023attentionneed} that generate translation directly from speech in the source language without relying on its transcription as an intermediate representation. It turns out that this approach is well suited for speech translation from spoken languages characterized by the lack of a standardized orthographic convention, which is the case for multiple low-resourced languages across the world including all Arabic dialects. In addition to being a way to get around the need of source language transcription, E2E models enables a simple and effective framework for transfer learning from pre-trained models on high-resource language pairs. In this work, we report our efforts to collect, annotate, and release the first open-source annotated Tunisian Arabic to English speech translation dataset. We also release a set of ready-to-use Speech Recognition and Speech Translation models alongside with a SpeechBrain recipe and the instructions needed to reproduce our results.\\

Our contributions are fourfold:

\begin{enumerate}
    \item \textbf{Data}: Release of TEDxTn, the first open source code-switching Tunisian to English speech translation corpus. 
    \item \textbf{Annotation quality}: Consistent and high-quality annotated corpus transcribed by professional transcribers. 
    \item \textbf{ASR and AST}: Development and evaluation of ASR and AST systems using multiple pre-trained Self-Supervised and multilingual models.
    \item \textbf{Open-Sourcing}: Data\footnote{Annotations released under a CC BY-NC-ND 4.0 license.}, annotation guidelines and models are released together with their code and training recipe\footnote{\url{https://huggingface.co/datasets/fbougares/TedxTn}}.
\end{enumerate}


\section{Related work}
Deep neural network approaches have revolutionized modern Natural Language Processing (NLP) tasks. However, these methods require large amounts of training data, which remain very limited for a large number of languages, including all Arabic dialects. Indeed, despite the considerable effort made to build datasets for multiple Arabic dialects, none of them could be considered today richly resourced. This is the case of all Arabic dialects where available speech datasets are, in general, scarce and even much scarcer when it comes to Code-Switching (CS) speech. CS speech processing has been gaining attention in recent years. This is particularly true for some languages, such as English-Mandarin \cite{6639094} \cite{Li13_interspeech} \cite{chiou-etal-2022-mandarin} or English-Hindi CS \cite{dey-fung-2014-hindi} \cite{DBLP:journals/corr/abs-1810-00662}. Previous works have studied Arabic speakers CS from linguistic and sociolinguistic perspectives \cite{Kais_Sultan_Mousa_Alowidha_2024} \cite{Abuhakema2013CodeSA}. Arabic speakers often switch from their dialectal Arabic to French or English\footnote{\scriptsize{There is one notable exception in Morocco, where some people use Spanish as CS language.}}. In fact, Arabic speakers generally code-switch to these two languages due to historical factors, since the Arab countries were mainly French and British colonies. Several studies investigated the reasons behind code-switching and pointed out that speakers generally switch for different reasons.
People can alternate languages in order to fill a lexical gap, when using more technical terms than native equivalents, to reflect modernity and sophistication, or when using foreign names without translation \citep{Takashi1990ASA}. In \citet{Eldin2014SocioLS}, the author studied the main drivers of Arabic-to-English switching among Facebook users and highlighted that incompetence, lack of facility, habitual expressions, and the speaker's mood are the main motivations behind CS. 

Although limited, there exist some previous works addressing CS in the domain of Arabic dialect Speech Recognition and Translation. In \citet{Elfahal2019AutomaticRA}, one hour of mixed Sudanese Arabic and English speech corpus was collected and recorded. Afterward, they used this corpus to train and evaluate a speech recognition system that achieved a 33\% word error rate (WER) on a test set of 25 sentences. 
A much larger amount of work was done to build ArzEn, a larger Egyptian Arabic and English CS corpus \citep{hamed-etal-2020-arzen}. ArzEn is a 12-hour corpus of mixed Egyptian Arabic-English speech. It is a collection of 38 recorded and transcribed interviews on broad topics, including education, hobbies, work, and life experiences. They achieved 57.9\% of WER \cite{hamed-etal-2022-arzen} using a CTC/attention-based end-to-end ASR system trained with the ESPnet toolkit \cite{watanabe18_interspeech}. This corpus was also extended to create ArzEn-ST \citep{hamed-etal-2022-arzen}, with translations into monolingual Egyptian Arabic and monolingual English. This a three-way speech translation corpus was used to train and evaluate various ASR, MT and AST systems. A multilingual strategy was proposed to model CS in Arabic speech recognition in \citet{Chowdhury2021TowardsOM}. They trained an E2E model using Arabic, English, and French data sets. Results are reported for Egyptian and Moroccan dialects. Although a low word error rate (WER) was reported for the Egyptian dialect CS ASR, a higher WER was observed for the Moroccan CS test set. Recently, a 48-hour Multi-dialectal Arabic Speech data set called Casablanca was collected and published in \citet{talafha-etal-2024-casablanca}. This data set aims to mitigate the data scarcity obstacle for a number of Arabic dialects. Casablanca covers eight Arabic dialects. It was used to evaluate various pre-trained SoTA multilingual speech models and fine-tuned Whisper-large-v2 models. We emphasize that only a subset of Casablanca is publicly available and does not include the Tunisian dialect. More details of the current literature on code-switched Arabic NLP are recently presented in \citet{hamedsurvey2025}.

With regard to the Tunisian dialect, the number of previous works related to ASR is still limited, and developed data sets are generally not available. Currently, there are only three publicly available ASR Tunisian dialect corpora, namely TARIC \cite{mdhaffar-etal-2024-taric}, TunSwitch \cite{abdallah2023leveragingdatacollectionunsupervised} and LinTo \cite{linagora2024Linto-tn}. TARIC is an 8-hour dataset that target the domain of human-to-human dialogues for train reservation tasks. Therefore, it was transcribed using only Arabic script. TunSwitch, on the other hand,  was collected from radio broadcasts that intentionally targeted the Tunisian Code-Switched ASR task \cite{abdallah2023leveragingdatacollectionunsupervised}. Overall, 8h15m of spontaneous Tunisian speech corpus has been collected as part of TunSwitch data set. This data set was used to train an end-to-end ASR system by fine-tuning the pre-trained speech encoder WavLM \cite{Graves2013SpeechRW} followed by three dense trainable layers trained with CTC loss \cite{Graves2013SpeechRW}.
Using a test set of about 25 minutes, the authors reported a WER of 29.47\% using an end-to-end ASR system and a 4-gram language model trained with an additional textual corpus of ten thousand monolingual English and French sentences. Recently, \cite{linagora2024Linto-tn} extended TunSwitch to create LinTo that contains an annotated data set of 81h38m. A kaldi \cite{Povey2011TheKS} based ASR system was trained using LinTo dataset, and a WER of 20.51\% was reported on the TunSwitch test set. In addition to the Tunisian dialect Speech resources mentioned above, there exists a data set used during the IWSLT \citep{anastasopoulos-etal-2022-findings} evaluation campaign and published recently in the Linguistic Data Consortium (LDC) catalogue. This data set represents 383 hours of manually transcribed conversational speech. A subpart of 160 hours of it is augmented with English translations. This data set was used by several teams within the context of IWSLT to develop multiple ASR and ST systems \citep{yan2022cmu} \citep{yang2022jhu} \citep{boito2022trac}. Although, the latter data set is adapted for Tunisian Arabic to English speech translation, we should point out that, unlike our data set, it is not publicly available and the input speech is conversational telephone recording sampled at 8Khz.

\section{Code-switching in Tunisian Arabic}
\label{section2}

Tunisia is the northernmost country in Africa. Its language is generally referred to as "Tunisian Dialect" or "Tunisian Arabic" or "Tounsi". Tunisia is an ethnically and linguistically homogeneous country, where 98\% of Tunisians identify as Arabs and speak Tunisian Dialect \cite{Ben_Youssef_Corpora2023}. Today's linguistic situation of Tunisia is strongly shaped by its history, trade, and today’s world. That's why Tunisian Arabic co-exists with Modern Standard Arabic (MSA) and French, in a ‘triglossic’ relationship\footnote{\scriptsize{We would like to highlight an increasing trend towards code-switching with English, compared to French, among Tunisian youth.}}. As a result of this situation, Tunisian daily communication is characterized by an alternation between multiple languages within a single conversation. 
This alternation between languages is commonly known as CS. It is defined as "the alternating use of two languages in the same stretch of discourse by a bilingual speaker." \cite{Bullock_Toribio_2009}. According to \citet{2013ContactLB} CS is at the same time a mechanism and an outcome of language contact. It is a significant and common linguistic phenomenon in Tunisia. It has been shown in \citet{Sayahi2011CodeswitchingAL} that the direction of the switch is almost always from Arabic to French, the most frequently switched categories are single nouns and noun phrases. With regard to CS frequency, the latter shows that education is the most important criterion. People with a higher education code-switch more compared to people with only a high school education. People with a university degree show a much higher frequency of CS, which reflects a higher degree of competence in the French language. However, gender does not affect the frequency of CS. Generally speaking, CS is studied at the sentence boundaries \cite{MyersScotton1989CodeswitchingWE,Poplack80} and classified into three types: inter-sentential, intra-sentential and extra-sentential switching. The following are descriptions of each type.
\begin{itemize}
    \item \textbf{Inter-sentential switching} defines the situation in which the alternation between languages occurs at sentence boundaries.
    \item \textbf{Intra-sentential switching}, on the other hand, refers to the alternation that occurs within the sentence without any indication of the shift.
    \item \textbf{Extra-sentential switching} also known as \textbf{tag-switching} is transplanting a tag from one language to another.
\end{itemize}

In addition to the above, there exists also the \textbf{intra-word switching}, where people change language within a single word occurs where Tunisian speakers attach Arabic clitics and affixes to foreign French or English words. Table \ref{tab1:examplesCS} provides a concrete example for each CS in the Tunisian dialect type extracted from the TEDxTN corpus.


\section{Corpus Creation}
\label{section3}

\subsection{Data Collection}
The source for this corpus is a collection of TEDx talks\footnote{\scriptsize{https://www.ted.com/about/programs-initiatives/tedx-program}}.
TEDx events are planned and coordinated independently. TEDx talks share the same format as TED talks. However, while TED talks are all in English, TEDx talks can be in a variety of languages, including local spoken languages and dialects. TEDx events aim to help communities, organizations and individuals produce TED-style events at the local level. They are planned and coordinated independently, on a community-by-community basis, under a free license from TED. 
TEDx talks are a valuable source for multiple speech processing tasks. They have been used to create many data sets for many languages and multiple tasks. Some examples of this are: (1)  TED-LIUM \cite{Hernandez_2018} created for English speech recognition; (2)  MTEDx \cite{salesky2021multilingualtedxcorpusspeech} built to support speech recognition and speech translation research across many languages and (3) TED-EL \citep{li-etal-2024-ted} created for Speech Entity Linking. TEDx talks are particularity a valuable source for speech processing of low-resource languages. However, they are usually the fruit of local non-funding initiatives. Therefore, the available recordings may be difficult to find on the Internet\footnote{Unlike TED talks, TEDx talks are not gathered in a common website and could be sometimes shared only on personal social network accounts} or of poor audio quality for speech processing tasks. Another notable difference between TEDx and TED events is the lack of volunteers who subtitle TEDx talks. 
In this version of the TEDxTN corpus, we were able to collect 108 talks with an acceptable audio quality ranging from 2 to 23 minutes. The audio quality of each TEDx talk was manually verified before saving it in WAV format sampled at 16Khz. Table \ref{tab0:general} shows some key statistics about the collected data.

\begin{center}
\begin{table}[ht]

\begin{tabular}{lc}
\hline
                     & \textbf{TEDxTN Corpus}    ~~~~~~ \\
\hline
\#Tedx Talks        &  108             \\
\#Tedx Events       &  38              \\
\#Different cities  &  11              \\
Languages            & TN/FR/EN        \\
Date range           & 2010 - 2023     \\
\#Speakers         &     130             \\ 
Total audio duration &  28h39min     \\ \hline
\end{tabular}
\caption{Overview of the TEDxTN corpus.}
\label{tab0:general}
\end{table}
\end{center}

\subsection{Corpus Annotation}
All TEDxTN talks were manually transcribed by professional tri-lingual (Arabic, English, and French) transcribers. Like all Arabic dialects, Tunisian Arabic does not have a standardized orthography. Therefore, words can have multiple correct spellings, and several letters can be used interchangeably. Moreover, in the context of CS speech, some loanwords are adapted and transformed by changing their pronunciation or integrating number, gender, or case agreement. All of that makes the definition and application of a unified transcription guideline particularly challenging. We have chosen to follow the CODA* \cite{Habash2018UnifiedGA} design principles to develop our annotation guidelines. Although CODA* included a seed lexicon, it remains limited and covers only five dialects (GLF, MOR, EGY, TUN, and LEV) with a very small lexicon for each dialect. In the context of this work, we derived a set of rules and patterns used to unify as much as possible the spelling for each annotator and between annotators. Below are some annotation rules extracted from our transcription and translation guidelines: 
\begin{enumerate}
    \item Use Arabic script for Arabic words and Roman script for foreign words.
    \item Use Arabic script for foreign words when they are adapted to Tunisian dialect. 
    \item Arabic clitics and affixes are written in Arabic script, and French or English words are written in Roman script. For example "\<هذا> point\<عال> ", "\textit{About this point}" in English.
    \item Use a predefined fixed spelling for common words like days of the week, numbers, quantities, percentages, etc.
    \item Negative pronouns are written attached such as \<مانيش>, "\textit{I am not}" in English.
    \item Translate to provide natural translations with the intended meaning rather than literal translations.
    \item Translate foreign words (i.e French) into fluent English while preserving the meaning present in the original code-switched text.
    \item Disfluencies such as partial words and repetitions should also be included in translations.
\end{enumerate}

In order to ensure a high-quality dataset, we followed a two-stage transcription process. The first stage takes as input the audio files and produces a segmented output with an initial transcription that may contain transcription errors or may also not be fully compliant with the transcription guidelines. The output of the first stage is systematically reviewed during a second validation stage, in which non-compliance with the guidelines and inattention errors are corrected. The English translation is performed using the Tunisian transcription with possible access to the corresponding audio recording in case of need. In most cases, we followed the LDC Arabic-to-English Translation Guidelines \citep{LDCGuidelines}.


\begin{table*}[htb]
\begin{tabular}{c|l}
\hline
\textbf{CS type }                      &   \textbf{TEDxTN samples} \\ \hline
\multicolumn{1}{c|}{\multirow{4}{*}{\parbox[m]{1mm}{\rotatebox[origin=c]{90}{\textbf{Extra-sent}}} }} & \hspace{5cm}   \<أذاكا لي خلاني إنزيد إندافع أكثر.>~~ \underline{Mais bon it happens}. \\
\multicolumn{1}{l|}{}                  &  \underline{Anyways, it happens}. This is what made me defend them even more. \vspace{1mm}  \\ \cline{2-2} 
\multicolumn{1}{l|}{}                  &   \hspace{4.9cm}.\<بالجواب هذاكا>  \underline{c'était un déclenchement d'amour} .\<حبتني> ~\\
\multicolumn{1}{l|}{}                  &   She loved me. \underline{Love was triggered} through this letter. \vspace{1mm} \\\hline\hline
\multicolumn{1}{c|}{\multirow{4}{*}{\parbox[m]{1mm}{\rotatebox[origin=c]{90}{\textbf{Intra-sent}}}}} &  \hspace{1.3cm} \<على رواحكم.>  \underline{five statements}  \<إذا كان نعطيكم تويكا نص ساعة ال كلكم تكتبوا لي> \\
\multicolumn{1}{l|}{}                  &   \underline{So}, when I saw it my heart started beating fast and I said to myself "Isn’t this it?" \vspace{1mm} \\ \cline{2-2} 
\multicolumn{1}{l|}{}                  &  \hspace{6.3cm}  .les bouquinistes\<هبطت لل>~ \underline{comme je suis} \<عنيد> \\
\multicolumn{1}{l|}{}                  &   Stubborn \underline{as I used to be}. I went to the booksellers. \vspace{1mm}  \\ \hline\hline
\multicolumn{1}{c|}{\multirow{4}{*}{\parbox[m]{1mm}{\rotatebox[origin=c]{90}{\textbf{Inter-word}}}}} & \hspace{2.6cm} .\<اللي عنده أذيكا ثروة> ~\underline{passion}\<وال> \<اللي عنده>~~~\underline{l'énergie}\<الشباب أذاكا وال> \\
\multicolumn{1}{l|}{}                  & These young people and their \underline{energy} and \underline{passion} are wealth.  \vspace{1mm} \\ \cline{2-2} 
\multicolumn{1}{l|}{} & \hspace{5.9cm}  \< كان عندي حلمة> \underline{préparatoire}\<وقت لي مشيت لل> \\ 
\multicolumn{1}{l|}{}                  &  When I started studying at the \underline{preparatory institute}, I had a dream. \vspace{1mm}  \\ \hline
\end{tabular}
\caption{Examples of different CS types in TEDxTN corpus followed by their English translation.
The underlining marks the non-Tunisian phrases and their corresponding translation in English.}
\label{tab1:examplesCS}%
\end{table*}

\subsection{Corpus Statistics}

In this section, we present an overview of the annotated TEDxTN corpus. Table \ref{codeswitch:tab1} includes the number of transcribed segments, words, and speakers. It also includes total speech duration, average segment length (words) and duration (seconds), as well as gender distribution. 

\begin{table}[H]
\begin{tabular}{@{}lll@{}}
\hline
  \textbf{Category} & \textbf{Value}  \\
\hline
\# Segments                                 & 17,278    \\
\# different speakers                       & 130     \\
Speech duration                             & 25h01min \\ 
Avg segment Duration  (seconds)             & 5.20 sec \\
Gender dist (M/F) - Count                   & 86/44 \\
Gender dist (M/F) - Duration                & 18h/07h \\
\hline
\#Total source words                             & 321,220 \\
\#Src TUN words                              & 177,079 \\
\#Src Intra CS words                          &   4,176 \\
\#Src foreign words                           & 43,932 \\
\#Seg. full Tun                            & 7,979\\
\#Seg. full foreign                        &  459\\
\#Seg. mixed                               & 9,299\\
\#Src Vocab size                           & 31,064 \\ \hline
\#Total target Words (Translation)            & 280,353  \\ 
\#Target Vocab size                           & 20,982  \\ \hline
\hline
\end{tabular}
\caption{Detailed statistics of TEDxTN corpus.}
\label{codeswitch:tab1}%
\end{table}

As reported in Table \ref{codeswitch:tab1}, we were able to transcribe around 25 hours of speech out of 28 hours and 39 minutes of audio signal (87.3\%). This represents about 17.2k segments containing more than 321k words from 133 different speakers and a vocabulary size of around 31k.

\subsection{Code switching statistics}

Only 7,963 segments out of 17,200 total segments are fully in Tunisian dialect. That means that 53.70\% of the TEDxTN-ST corpus segments contain at least one foreign word. In order to better quantify the amount of code-switching present in TEDxTN-ST data, we calculate the Code-Mixing Index (CMI). CMI was introduced by \citet{das-gamback-2014-identifying} as a method to compare different code-mixed corpora to each other. CMI is defined as: 
\begin{equation}
    CMI = \frac{\sum_{i=1}^{N} w_{i} - max\{w_{i}\}}{ n - u}  
\end{equation}
where $\sum_{i=1}^{N} w_{i}$ is the total number of words from $N$ languages, $w_{i}$ is the number of words in language $i$, $n$ is the total number of words regardless of language, and $u$ is the number of tokens given language-independent tags. CMI is equal to 0 for utterances that contain only tokens from one language. A high CMI score is an indicator of the high degree of code-mixing in the text. The CMI for the entire TEDxTN corpus is 21.50\%. This indicates a high rate of CS in this corpus. As shown in Table \ref{codeswitch:tab1}, we also include statistics on the number of Tunisian words, written in Arabic script, (\textbf{Src. w TUN}), the number of foreign words fully spelled in Latin script (\textbf{Src w. foreign})  and the number of words written using a mix of Arabic and Latin script (a.k.a intra-word switching). On the word level, among code-mixed sentences (the 9.299 sentences reported in \textbf{\#Seg.~mixed}  row), 67.78\% of the words are Arabic, 26.38\% are foreign, and 5.84\% are intra-words code-switch. 

\subsection{Trigger Words}

As defined in \citet{hamed-etal-2018-collection}, code-switching trigger words are words that can prompt a bilingual speaker to switch languages during a conversation. TEDxTN includes 2729 unique Arabic code-switching trigger words.

\begin{table}[H]
\centering
\begin{tabular}{@{}llll@{}}
\hline
 \textbf{Word}   & \textbf{English} & \textbf{Frequency} \\\hline
\<ال> & The & 2,688 \\
\<آ - أ> & Hesitation  &  988 / 225 \\
\<و> & And & 436 \\
\<في> & In &  399 \\
\<لل> & To &  247 \\
\<متاع> & Belongs to  & 200 \\ 
\<بال> & With   &  173\\ 
\<معنتها> & This means   &  129\\

\end{tabular}
\caption{TEDxTN most frequent trigger words.}
\label{trigger}%
\end{table}

 Table \ref{trigger} shows the most common trigger words that precede a code switching point in TEDxTN. The most common switches occur after the definite article \<ال> (The). This is reasonable because \<ال> is placed before a foreign noun or adjective that the speaker wants to specify. The \<و> and \<في> trigger words are aligned with the observations reported for the Egyptian dialect in \citet{hamed-etal-2018-collection}. As for \<متاع>  and \<معنتها>, they are very common transition words used in Tunisian dialect.

\section{Experiments and results}

\subsection{Data split}
\label{sec:split}
We created standardized data splits for training, validation, and evaluation. We have chosen to put full TEDx talks in dev and test sets in order to avoid contamination between the training and evaluation. The number of talks, segments, and words are reported in Table \ref{split:tab1}. We also report the total duration, the number of unique speakers, the gender distribution, and the CMI score per dataset. As shown in Table \ref{split:tab1}, speakers belonging to the validation and test subsets are not seen in the training set. In addition, validation and test sets have higher CMI scores compared to training data. Finally, we also ensured that both male and female speakers are kept within the validation and test set.

\begin{table}[htb]

\begin{tabular}{@{}lllll@{}}
\hline
                & \textbf{Train}   & \textbf{Valid.} & \textbf{Test} \\\hline
\#Talks         &  97     & 05     & 06     \\
\#Segments     &  15,626 & 731    & 842    \\
\#Words        &  205,753& 11,250 & 11,834 \\
Duration        &  22h40m & 01h07m & 01h14m \\
\#Speakers      &  117    & 10     & 07     \\ 
CMI score       &   20.66 & 24.37  & 33.09  \\
Gender: M/F     & 77/40  & 5/5    & 5/2    \\ \hline
\end{tabular}
\caption{TEDxTN corpus split to train, valid and test.}
\label{split:tab1}%
\end{table}

\subsection{Automatic Speech Recognition}

\begin{table*}[ht]
\centering

\begin{tabular}{lccccc}
\hline
\textbf{Model}   & \textbf{Model Size} & \multicolumn{2}{c}{\textbf{Valid.}}  & \multicolumn{2}{c}{\textbf{Test}} \\
               & {(\#Params)}      & WER ($\downarrow$) & CER ($\downarrow$)    & WER ($\downarrow$) & CER ($\downarrow$) \\
\hline
Whisper-small  (zero-shot)    & 244M   &      133.24     &     100.00 & 183.81 &  142.00  \\
Whisper-meduim  (zero-shot)   & 769M   &      130.61         &     103.00  & 150.71  & 122.00  \\
Whisper-Lg-v3 (zero-shot)     & 1550M  &  92.50&63.20    &  94.00 &67.90  \\ 
\hline \hline
Whisper-Small   & ~244 M   & 26.66&11.81    &  27.78&13.38  \\ 
Whisper-Medium     & ~769 M &  23.10 & 10.46   &  25.37 &13.00  \\  
Whisper-Lg-v3      & 1550 M  &  22.72&09.77    &   25.19&12.33  \\ \hline
MMS Large          & 316.6 M &  35.43&13.02    &  37.29&14.67  \\
MMS 1B             & 964.3 M &  26.78&09.90    &  27.91&11.27  \\ \hline
XLSR Large         & 316.6 M &   35.74&13.79   &  37.11&15.26  \\
XLSR 1B            & 964.3 M &  28.12&10.82    &   29.98&12.12  \\ \hline
w2v-Bert-2.0T    & 590.1 M & \textbf{19.92} & \textbf{07.10}    &  \textbf{21.37}& \textbf{08.34}\\ \hline
\end{tabular}
\caption{ASR results of TEDxTN Tunisian Arabic speech. Lower WER and CER indicate better quality.}
\label{tab:ASR:results}%
\end{table*}

Given the relatively small size of our datasets, we opt for a fine-tuning approach rather than training a Tunisian dialect ASR system from scratch. As regards the choice of the pre-trained models to use, there are various options available to us, ranging from small models with a few hundred million parameters to bigger models with around 1 billion parameters. In addition to the model size, we also have the choice between multiple model architectures. In this work, we experimented with fine-tuning 5 different pre-trained models. Namely, we use the TEDxTN training set to adapt Whisper \cite{whisper}, Massively Multilingual Speech (MMS) \cite{mms}, XLSR \cite{xlsr} and w2v-Bert-2.0T \cite{seamless} models. All of our experiments were performed using the SpeechBrain toolkit \cite{speechbrain} without using a language model. All our models were trained for 80 epochs. For whisper based models, we used the original encoder-decoder architecture without any parameters freezing. MMS and XLSR models are trained using an additional linear layer of size 1024 followed by a Connectionist Temporal Classification layer (CTC) for transcribing the labels. Finally, for W2v-Bert-2.0T model we added two transformer layers of size 1024 each, followed by a CTC layer for transcribing the labels. We used Adam optimizer for all our ASR models.  Our results are reported in Table \ref{tab:ASR:results}. We evaluated Whisper (large-v3), one of the best multilingual open source speech recognition models, in a zero shot setting (line 1 in Table 6) and we observed, as shown in previous work for other Arabic dialects \cite{talafha2023}, that Whisper did not reach reasonable performance with 92.50\% and 94.00\% WER on TEDxTN dev and test sets, respectively.\\
Fine-tuning Whisper models using domain-specific data results in a significant reduction in WER. We started by fine-tuning Whisper-small, which already gives a significant WER reduction compared to a much bigger model (large-v3) in a zero-shot setting. Using larger Whisper models (large-v3) incrementally decreases the WER to achieve \textbf{25.19\%} WER on the test set. We also report obtained results when fine-tuning MMS Large and MMS 1B models. As shown in the table, MMS 1B obtained better results compared to MMS Large. However, MMS 1B results are comparable to Whisper-small, although the latter has about 4 times fewer parameters. An interesting observation is that fine-tuning the w2v-Bert-2.0T model gives much better results compared to Whisper-large-v3 while having 3 times fewer parameters. w2v-Bert-2.0T model achieves \textbf{19.92\%} and \textbf{21.37\%} WER on TEDxTN dev and test sets, respectively.

       

\subsection{Automatic Speech Translation}

For the same reasons set out above, we decided to opt for a fine-tuning approach of pretrained models. We started by using pre-trained Speech translation models. Particularly speaking, we began by translating the dev and test set in a zero shot fashion using different pre-trained Whisper models. Next, we fine-tuned these models using TEDxTN dataset. To be consistent with the ASR experiments, we kept the same data split used reported in section \ref{sec:split}. All our translation outputs are evaluated with TrueCased BLEU score without punctuation using sacrebleu \citep{post-2018-call}. Table \ref{AST-results} shows the speech translation performance of each trained model. All our models are fine-tuned for 80 epochs following the default Whisper recipe of SpeechBrain toolkit.

\begin{table}[htb]
  \centering
  \begin{tabular}{lll}
    \hline
    \textbf{Model}  & \textbf{Valid.} ($\uparrow$) & \textbf{Test} ($\uparrow$) \\ \hline
       Whisper-small \footnotesize{(zero-shot)}    &  3.98  &  5.70\\
       Whisper-med. \footnotesize{(zero-shot)}       & 10.23  & 12.85 \\ 
       Whisper-lg-v3 \footnotesize{(zero-shot)}     & 10.96  & 13.95 \\ \hline
       Whisper small        & 17.31 & 18.53 \\
       Whisper med.         &  23.02    & 24.50 \\
       Whisper-lg-v3          &  \textbf{25.19}    & \textbf{25.68} \\
    \hline
  \end{tabular}
  \caption{BLEU scores of TEDxTN Speech Translation.}
  \label{AST-results}
\end{table}

As expected, we obtained better BLEU score using larger Whisper models for both zero-shot and fine-tuning settings. For instance, the best zero-shot BLEU scores are obtained using whisper-Large-v3 (row whisper-Lg-v3) with 10.96 and 13.95 for valid and test respectively. Likewise, whisper-Large-v3 achieves the best performing model after fine-tuning on TEDxTN training set with \textbf{25.19} and \textbf{25.68} BLEU scores for validation and test sets respectively. Note that we trained also speech translation models by feeding WLSR and w2v-Bert-2.0T encoders outputs to the NLLB decoder, in its 1.3B parameters configuration. Contrary to what we expected, the model did not in exceed a BLEU score of 5. More investigation of this model is left to be done in future work.

\begin{table*}[ht]
\centering
\begin{tabular}{l|l}
\hline
    \textbf{Output}   & \textbf{TEDxTN Samples} \\
\hline
Reference & \<ما  يحبش  يبدل> \textbf{Ooredoo} \<متاعه في> Il il est près à passer le reste de jours \\
Prediction & \hspace{2mm}  \<ما  يحبش  يبدل> \< متاعه في \textbf{أوريدو}> Il il est près à passer le reste de jours \\
\hline \hline 
  Reference  &  \< إنحب نقول له حاجة> \textbf{fumoir}+\<آنا ثمة شكون ما يتكيفش ويقعد في ال>  \\
Prediction   &\hspace{4mm}  \<آنا ثمة شكون ما يتكيفش ويقعد في \textbf{الفيموار}  إنحب نقول له حاجة>  \\

\hline \hline 
  Reference  &\hspace{3.5cm}  \<تي حتى \textbf{مالسيركيلاسيون}  ولات أمورها واضحة>
\\ \hline
Prediction   &\hspace{3.5cm} \<ولات أمورها واضحة> \textbf{circulation} \<تي حتى مال>  \\ 

\hline \hline
  Reference    & \hspace{1.8cm} \textbf{chef} \<متاع تونس عال> mentalité \<وحدك تنجم تبدل ال> à toi \\
  Prediction   & \hspace{1.8cm} \<متاع تونس \textbf{عالشاف}> mentalité \<وحدك تنجم تبدل ال> à toi  \\ \hline \hline

\end{tabular}
\caption{Examples of ASR (w2v-Bert-2.0T model) errors from TEDxTN test set.}
\label{tab1:err-examples}%
\end{table*}

\subsection{Qualitative Analysis}

\textbf{Speech transcription:} To understand the quality of ASR transcription per segment type, we divided the test set into the following 3 subsets: (1) \textbf{TUN} subset with segments uttered only in the Tunisian dialect, (2) \textbf{MIXED} subset includes code-switching segments and (3) \textbf{FOR} subset with segments fully in foreign language. Using our best ASR system (w2v-Bert-2.0T from Table \ref{tab:ASR:results}), we calculated the WER for each subset. 

\begin{table}[htb]

\begin{tabular}{lcccc}
\hline
            & \textbf{TUN}  & \textbf{MIXED} & \textbf{FOR} & \textbf{ALL}\\\hline
\#Seg.  &  215   & 551     & 76  & 842 \\ \hline
\#Words     &  1,912 & 9,282   & 639  & 11,833\\ \hline
WER ($\downarrow$) &  24.16  &  21.31   & 13.93 &  21.37 \\ \hline
\end{tabular}
\caption{ASR Error analysis per segment type.}
\label{analysis:tab1}%
\end{table}

As shown in Table \ref{analysis:tab1}, most ASR errors are made for Tunisian-only and code-switched segments. Manual inspection of the code-switched segments shows that the system outputs the correct transcription but using different script from the one used in the reference. Some examples of this are provided in Table \ref{tab1:err-examples}. In the first example, our ASR system transcribed the word  \<الفيموار> in Arabic while this word is written using the prefix \<ال> plus the same word in Latin script "fumoir". Same for the word "circulation", but in the opposite direction: The reference is written in Arabic script (\<مالسيركيلاسيون>) while the human transcription is in Latin script. As regards, speech translation system, we analyzed the output of the fine-tuned Whisper large-v3 model but no particular error pattern was identified. 

\section{Conclusion and future works}

In this study, we propose TEDxTN, the first Tunisian Arabic to English Code-switching Speech Translation annotated corpus. \textbf{TEDxTn} is carefully annotated by linguistic experts following a detailed annotation guideline. This corpus was used to train and evaluate multiple strng Tunisian dialect speech transcription and translation baselines. Our best models achieves \textbf{21.37\%} WER and \textbf{25.68} BLEU scores on the transcription and translation tasks of TEDxTn test set respectively. We believe that this corpus fills an important resource gap in Code-switching research for Tunisian dialect. For future work, we plan to extend \textbf{TEDxTn} as new talks are available and use it for other NLP tasks.

\section*{Ethical considerations and limitations}

Like any other dataset, the collected speech corpus is not representative of all the spoken forms of Tunisian Dialect. This corpus is likely unbalanced in terms of any demographic aspect since it includes talks from only 11 different cities in Tunisia. Nevertheless, we think that the lack of previous code-switching speech Tunisian Arabic to English translation data set, would make it valuable resource for training and evaluating code-switching speech models of Tunisian Dialect.
TEDx talks are governed by the CC BY-NC-ND 4.0 license. Under this license, “NoDerivatives” implies that any modifications, remixes, or transformations cannot be distributed. In compliance with this we distribute only transcriptions and translations. For the audio recordings, we provide the YouTube URL of each video for users to download.

\section*{Acknowledgments}

This work was performed using HPC resources from GENCI-IDRIS (grant AD011015051R1) and received funding from the ESPERANTO research and innovation programme under the Marie Skłodowska-Curie (grant No 101007666).





\bibliography{custom}

\end{document}